\documentclass[letterpaper, 10 pt, conference]{ieeeconf}  

\IEEEoverridecommandlockouts                              

\usepackage{graphicx} 
\usepackage{amsmath} 
\usepackage{amssymb}  
\usepackage{subcaption}
\usepackage{color}
\usepackage{booktabs}

\title{\LARGE \bf
Pre-trained Visual Representations Generalize Where it Matters in Model-Based Reinforcement Learning}

\author{Scott Jones$^{1}$, Liyou Zhou$^{1}$ and Sebastian W. Pattinson$^{1}$
\thanks{$^{1}$ Institute for Manufacturing, Department of Engineering, University of Cambridge}%
}

\begin{document}

\maketitle
\thispagestyle{empty}
\pagestyle{empty}

\begin{abstract}
In visuomotor policy learning, the control policy for the robotic agent is derived directly from visual inputs. The typical approach, where a policy and vision encoder are trained jointly from scratch, generalizes poorly to novel visual scene changes. Using pre-trained vision models (PVMs) to inform a policy network improves robustness in model-free reinforcement learning (MFRL). Recent developments in Model-based reinforcement learning (MBRL) suggest that MBRL is more sample-efficient than MFRL. However, counterintuitively, existing work has found PVMs to be ineffective in MBRL. Here, we investigate PVM's effectiveness in MBRL, specifically on generalization under visual domain shifts. We show that, in scenarios with severe shifts, PVMs perform much better than a baseline model trained from scratch. We further investigate the effects of varying levels of fine-tuning of PVMs. Our results show that partial fine-tuning can maintain the highest average task performance under the most extreme distribution shifts. Our results demonstrate that PVMs are highly successful in promoting robustness in visual policy learning, providing compelling evidence for their wider adoption in model-based robotic learning applications.
\end{abstract}

\section{INTRODUCTION}
Vision is a critical sensing modality for enabling robots to operate in complex, unstructured environments. Reinforcement learning (RL) has emerged as a powerful framework for training agents directly from raw visual input, capable of acquiring sophisticated behaviors from simple reward signals \cite{sutton}. Classical methods in visual RL train both a visual encoder and a policy from scratch \cite{e2e-bottle}. This approach is notoriously data-hungry, often requiring hundreds of millions of steps of task experience \cite{openai-rubik}. More importantly, policies trained in this manner tend to generalize poorly when faced with out-of-distribution (OOD) inputs in its deployed environment \cite{e2e-bottle, generalise-coin}. Consider an autonomous driving example, adverse weather or change in street lighting can easily produce novel visual scenarios unseen in training. Out-of-distribution (OOD) generalization in such scenarios is imperative for the safe and successful operation of robotic systems.

A promising approach to OOD generalization is to leverage representations from pre-trained vision models (PVMs) to encode observations. Transferring pre-trained representations is well established in computer vision for increasing training efficiency, outperforming models trained from scratch \cite{dinov2, CLIP}. Research into the paradigms of model-free reinforcement learning and imitation learning (IL) has shown that using PVMs improves policy generalization and learning speed \cite{unsurprising, vc-1}. In contrast, Model-based reinforcement learning (MBRL) has received far less investigation, despite often being more efficient \cite{complexity-mbrl} and robust to distribution shifts \cite{benefits-mbrl} than MFRL. To date, only a single study has investigated the integration of PVMs with MBRL\cite{surprising}. Their findings were counterintuitive: neither sample efficiency nor generalization improved.  Their analysis, conducted in simplified environments with limited realism and modest distributional shifts (e.g., color perturbations), attributed this outcome to the fixed nature of frozen pre-trained representations, which constrained the reward modeling capacity of the world model and hindered generalization.

\begin{figure}
    \centering
    \includegraphics[width=\columnwidth]{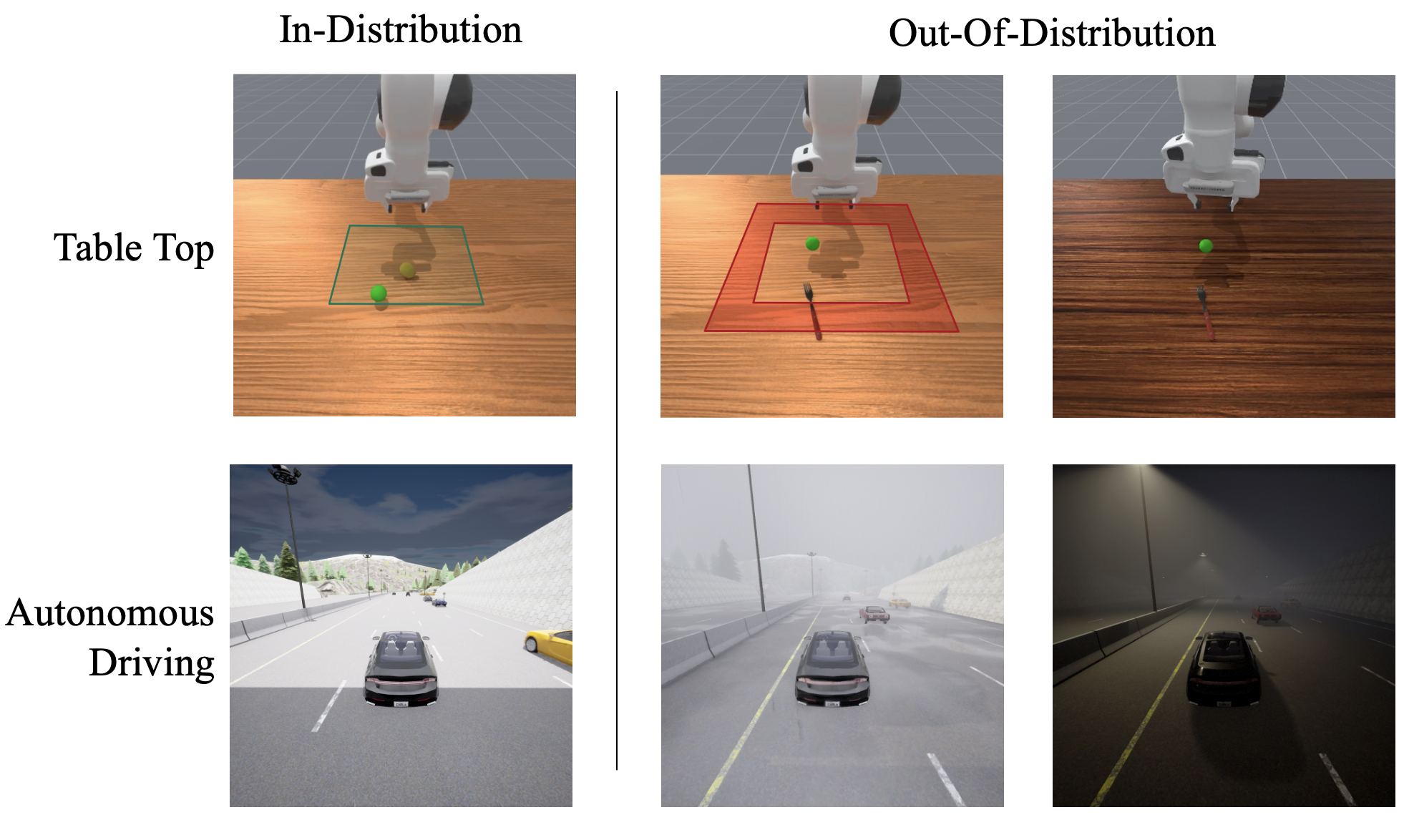}
    \caption{The ID and OOD visualization of our environments. In the table top task (top), the ID objects spawn in the green square, and the OOD objects spawn in the red ring. The hard shift changes to a different table texture. In the autonomous driving task (bottom), the weak and strong OOD weather shifts get progressively more foggy, rainy, and dark.}
    \label{fig:envs}
\end{figure}

These findings provide strong motivation to further investigate the topic, given the advantages of MBRL and the positive results using PVMs in other policy learning paradigms. The key contributions of our work are as follows:

\begin{enumerate}
    \item \textbf{Benchmarking \textit{hard} OOD performance of MBRL with PVMs}: we extend the experiments of \cite{surprising}, evaluating policies on more challenging and realistic visual tasks. To this end, we introduce the concept of \textit{hard} distribution shift, where aspects of the task that remained constant during training are altered for evaluation. Our findings show that under such shifts, MBRL agents using PVMs can generalize far better than learning from scratch.
    
    \item \textbf{Benchmarking MBRL with Varying Degrees of PVM Fine-tuning}: To address the proposed downsides of fixed representations, we also investigate the effects of fine-tuning PVMs end-to-end. PVMs are evaluated in three configurations: frozen weight, partial fine-tuning, where select layers are updated, and full fine-tuning. We find the partial fine-tuning approach presented the strongest combination of in-distribution (ID) and OOD performance.
    
    \item \textbf{Analysis of PVM Properties for Generalization}: We analyze several properties of the PVMs and agents to uncover why certain configurations are more robust to task distribution shifts. Our findings indicate the vision models that were invariant to input perturbations and had less forgetting tended to have the strongest agent generalization.
\end{enumerate}

\section{BACKGROUND}

\subsection{Partially Observable Markov Decision Processes}
For visual policy learning, we model both problems as Partially Observable Markov Decision Processes (POMDPs) \cite{partial-mdp} since state information must be extracted from image observations. A belief state $b(s)$ is maintained, a distribution over the state space $S$. States transition with probability $T(s_{t+1} | a_t, s_t)$ given an action $a_t$. An observation of the new state $o_{t+1}$ is used to update the belief state and inform a learned policy to sample the next action $a_{t+1}\sim\pi(b_{t+1})$.

\subsection{Model-Based Reinforcement Learning}

In MBRL, a model is trained to predict the transition probabilities and reward, given a state-action pair \cite{sutton}, whereas MFRL implicitly learns these through trial-and-error interactions in the environment. Under certain conditions, MBRL can be exponentially more efficient than MFRL \cite{complexity-mbrl}. Moreover, it has been shown in theory and experiments that MBRL policies can also generalize better. For environments with factored structure, where the environment's rules $T(s_{t+1}|a_t, s_t)$ can be decomposed, the world modeling constraint removes implausible Q-functions that do not obey the environment dynamics, leading to better generalization \cite{benefits-mbrl}.

\subsection{DreamerV3}
We investigate the SOTA MRBL algorithm DreamerV3 that was also used in related work \cite{surprising}. In the DreamerV3 architecture, there are two distinct learning loops. The world model learns to predict the dynamics of the environment. The actor and critic networks learn a policy to maximize returns \cite{dreamerv3}. 

\subsubsection{World Model}
The world model is implemented as a Recurrent State-Space Model (RSSM) consisting of a CNN encoder and decoder, and a gated recurrent unit (GRU). The encoder takes visual observations $x_t$ and produces a categorical distribution $q_{\phi}(z_t | h_t, x_t)$ from which a discrete stochastic representation $z_t$ is sampled. The world state $s_t$ is the concatenation of the GRU's recurrent state $h_t$ and $z_t$. 

At each time step, the GRU updates the hidden state, given an action and world state. The MLP dynamics model predicts the future distribution $p_{\phi}(z_{t+1} |h_{t+1})$. The encoder produces the posterior representation $q_{\phi}(z_{t+1} | h_{t+1}, x_{t+1})$ upon a new observation. The dynamics model learns to match the encoder output without needing observations, which enables the imagined rollouts used to train the actor and critic networks. Dynamic prediction loss is calculated as the KL divergence between the two distributions over $z_{t+1}$. The decoder reconstructs observations from the predicted future world state $s_{t+1}$ to compare with the actual next observation $x_{t+1}$. Training also involves a reward predictor for the instantaneous reward of each state and a continue predictor for terminal states. 

\subsubsection{Actor-Critic Learning}
The actor and critic models are provided the world model state $s_t$ as input. The critic network learns to approximate the distribution of returns for each state under the current actor's policy. With predicted values $v_t$ being the expected value under the distribution $v_{\psi}(\cdot|s_t)$. The critic learns based on the maximum likelihood loss from the bootstrapped lambda returns \cite{sutton}. The actor network's role is to select actions that maximize the return. It generates a distribution over actions known as its policy $\pi_\theta(a_t|s_t)$. The actor network is updated using a version of the REINFORCE algorithm with entropy regularization to promote exploration.

\section{RELATED WORK}
\subsection{Visual Policy Learning for Control}
Classical approaches in this field often follow the \textit{tabula rasa} paradigm, where both a policy and a visual encoding are learned from scratch \cite{e2e-bottle, openai-rubik}. The generalization of learned policies is often weak. \cite{e2e-bottle} show that success rates fall dramatically when distractor objects are present in the frame. Agents often need to be presented with millions of different task permutations in training before learning to generalize \cite{generalise-coin}. Moreover, training visual policy learning can be very sample inefficient. For example, training a robot hand using vision to solve a Rubik's cube took several months and 64 GPUs, amounting to a total of 13K years of experience \cite{openai-rubik}. 

\subsection{Pre-trained Visual Representations in Policy Learning}

Many studies have experimented with leveraging PVM in policy learning to improve generalization. \cite{unbiased-look} investigated the choice of dataset for vision model pre-training. Using imitation learning using masked autoencoders (MAEs) pre-trained on different datasets they showed that policies using classical image datasets, such as ImageNet, generalize well to novel objects and positions. Whereas, the tabula rasa model averaged a 0\% success rate across all three tasks tested. Using PVMs often improves sample efficiency as the policy network can immediately leverage information-rich representations of the environment. For example, VC-1 is a vision model pre-trained on a mixture of egocentric and classical image datasets to be applied to visual control tasks \cite{vc-1}. Given the same training budget, the PVM-based agents had a greater success rate than a tabula rasa approach. These results have been consistently found for both MFRL and IL. For brevity, we summarize their findings in Table \ref{tab:pvr-pl-summary}, covering the policy learning paradigms investigated and whether using PVMs improved efficiency or generalization.

Fewer papers have investigated the use of PVMs in MBRL. \cite{surprising} integrated popular PVMs, including CLIP \cite{CLIP} and DINOv2 \cite{dinov2} with frozen weights into two MBRL algorithms: DreamerV3 \cite{dreamerv3} and TD-MPC2 \cite{tdmpc}. Agents were evaluated in modified environments with distributional shifts, such as changes in object color and position. For training, the agents were exposed to 80\% of such changes, and the remaining 20\% was used for evaluation. They found that sample efficiency did not improve when using PVMs, and the OOD generalization was actually worse than the baseline tabula rasa approach. The paper posits that learning speed did not improve since the decoupled objectives of MBRL make it harder for the agent to adapt to the fixed representations of PVMs compared to those learned from scratch. They also find that the baseline representations were able to better separate high and low reward observations than the \textit{frozen} PVMs and identified a significant correlation between world model reward prediction and OOD generalization. 

Such findings provide strong motivation for further research. Firstly, we identify fine-tuning our PVMs as a solution to the downsides of fixed representations presented in their study. Moreover, in their experiment, the distribution shifts were mild, and the agent was exposed to shifts of the same type during training. In the real-world deployment, we cannot practically anticipate all the types of shifts an agent will experience when deployed. This motivates the need to test OOD with unseen shift types, which we denominate as \textit{hard} shifts. Results in these conditions should better reflect the real-world generalization of policies, \textit{where it matters}. 

\begin{table}[]
\centering
\begin{tabular}{lcccc}
\textbf{Paper}                & \textbf{Policy} & \textbf{Sample Efficiency}   & \textbf{Generalization} \\
\midrule
\cite{unsurprising}           & IL         & $\color{green}{\uparrow}$   & -                             \\
\cite{r3m}                    & IL         & $\color{green}{\uparrow}$   & -                             \\
\cite{ve-gaming}              & IL         & =                           & -                             \\
\cite{what-makes-pvrs}        & IL         & -                           & $\color{green}{\uparrow}$     \\
\cite{unbiased-look}          & IL         & -                           & $\color{green}{\uparrow}$     \\
\cite{vc-1}                   & IL \& MFRL & $\color{green}{\uparrow}$   & -                             \\
\cite{offline-visual}         & IL \& MFRL & -                           & -                             \\
\cite{pvrs-rl}                & MFRL       & -                           & -                             \\
\cite{mvp}                    & MFRL       & $\color{green}{\uparrow}$   & $\color{green}{\uparrow}$     \\
\cite{robust-mid}             & MFRL       & $\color{green}{\uparrow}$   & $\color{green}{\uparrow}$     \\
\cite{role-pvrs}              & MFRL       & -                           & $\color{green}{\uparrow}$     \\
\cite{pie-g}                  & MFRL       & $\color{green}{\uparrow}$   & $\color{green}{\uparrow}$     \\
\cite{mid-lvl-gen-eff}        & MFRL       & $\color{green}{\uparrow}$   & $\color{green}{\uparrow}$     \\
\midrule
\cite{surprising}             & MBRL       & =                           & $\color{red}{\downarrow}$     \\
\bottomrule
\end{tabular}
\caption{Findings from papers investigating PVMs in visual policy learning. Results are presented symbolically as follows: improvement $\color{green}{\uparrow}$, no change $=$ and worsening $\color{red}{\downarrow}$. Papers that do not investigate an outcome are signified by -.}
\label{tab:pvr-pl-summary}
\end{table}

\section{METHODS}
\subsection{Modifying DreamerV3}  
In our work, we choose to modify the default 200M parameter DreamerV3 model. We target the CNN encoder, which processes visual input $x_t$ into a latent representation $e_t$, which is then projected down by a feed-forward layer to output the distribution over $z_t$. Most PVMs have excellent downstream performance adapted using a simple linear layer \cite{dinov2}, which is already present in the DreamerV3 architecture. Thus, only the CNN is replaced, and the rest of the architecture remains precisely the same. 

In our experiments, vision models are evaluated under degrees of fine-tuning: frozen, partial fine-tuning, and full fine-tuning. We investigate partial fine-tuning the PVMs since it can reduce forgetting \cite{half-ft-llm} and improve generalization compared to fully fine-tuning \cite{peft-vs-ft}. We fine-tune only the last quarter of layers, which has been shown to reach 90\% of the performance of full fine-tuning for transformer models \cite{ww-elsa-do}. The fine-tuning learning rate is set to $4\times10^{-6}$, a $\frac{1}{10}$ of the learning rate of the rest of the system.

\subsection{Vision Models}
Our baseline model is the DreamerV3 CNN trained from scratch. Consisting of repeated units of convolutional layers and max pooling, the feature map dimensions are halved at each pass. To match the setup of \cite{surprising}, we increase image resolution from $64\times64$ to $128\times 128$ for the baseline and $224\times 224$ for the PVMs. We add another convolutional and pooling unit, preserving the output dimensions of the original DreamerV3 CNN. As the reconstructed output is higher resolution, we add another upscaling layer to the decoder for both baseline and PVMs.

The two PVMs we investigated were DINOv2 and CLIP. Our selection criteria were guided by proven effectiveness in visual policy learning \cite{pvrs-rl, ve-gaming, cliport} and compatibility with the DreamerV3 JAX implementation. For memory efficiency, we integrated the smallest available PVMs. For DINOv2, following \cite{dino-wm}, we select $e_t$ to be the patch embeddings, that is $e_t \in \mathbb{R}^{N\times E}$, where $N$ is the number of patches and $E$ is the embedding dimension (384 for ViT-S). For CLIP, $e_t$ is the output for the \texttt{[CLS]} token, following \cite{vc-1}.

Table \ref{tab:model_comparison} summarizes our vision models. While the embedding sizes differ, we compare encoder performance under the same information budget, as outputs are projected down to the same size representation $z_t$. Unfortunately, without pre-training our own models, model size is a variable that we cannot control. Although the findings of \cite{surprising} do not appear to favor increased capacity.

\begin{table}[]
\centering
\begin{tabular}{lccc}
\textbf{Model} & \textbf{Pre-training Dataset} & \textbf{\# Parameters} & \textbf{Embedding Size} \\
\midrule
Baseline & N/A  & 5.1M & $4\times4\times256$\\
\midrule
DINOv2   & 142M & 21M & $256\times384$ \\
CLIP     & 400M & 88M & 768 \\
\bottomrule
\end{tabular}
\caption{Comparing differences between our tested vision models.}
\label{tab:model_comparison}
\end{table}

\subsection{Environments}
To test policy generalization, our approach is most similar to \cite{rl-vigen}, where agent generalization is evaluated across varying degrees of shift. We strive to test in environments with stronger realism, both to increase visual complexity and closer approximate real-world performance. As such, two new environments (Fig. \ref{fig:envs}) were implemented for our use case:

\subsubsection{Table Top Environment}

We use the ManiSkill simulator \cite{maniskill} to create a table-top manipulation environment with distribution shifts. We replicate the setup designed by \cite{surprising} on the tabletop manipulation \texttt{pick\_single\_ycb} task. In each episode, an object is sampled from the YCB dataset and placed at a random location. The robot arm, using a wrist-mounted camera, must pick up the object and bring it to a goal location, signified by a green sphere. 

\begin{itemize}
    \item \textbf{ID setup}: 20\% of the YCB objects and positions are held out from the training set. We modify the original task so objects only spawn within a tighter square, 80\% of the original area ($0.2\text{m} \rightarrow 0.1789\text{m}$).
    
    \item \textbf{Easy OOD setup}: Objects are sampled from the 20\% unseen set. Objects spawn in an outer ring, the remaining 20\% area of the original square.
    
    \item \textbf{Hard OOD setup}: On top of the easy shift, a new table texture is introduced. We change from a light to a darker wood texture, a realistic visual shift.
\end{itemize}

\subsubsection{Autonomous Driving Environment}

The RL-ViGen \cite{rl-vigen} benchmark suite is used to create distribution shifts in an autonomous driving scenario. RL-ViGen is a benchmark for testing visual RL generalization. It aims to increase realism and difficulty in standard robot simulation environments by implementing adjustable shifts in lighting, scene structure, and views. We select their autonomous driving CARLA benchmark as it has the greatest realism and dynamics complexity. In this task, the ego vehicle aims to drive on a four-lane highway and is surrounded by 40 other randomly spawned vehicles, using a camera view from the front of the vehicle. The reward function encourages driving distance while penalizing collisions and unnecessary steering. The RL-ViGen evaluation environments are split into difficulty levels: weak, medium, and strong. The main changes between difficulty levels are increasingly adverse weather and lighting conditions. We note that all evaluation settings would be considered \textit{hard} shifts, as such lighting and weather scenarios are unrepresented in training.

\begin{itemize}
    \item \textbf{ID setup}: Every episode, the ego vehicle starts at the same location on the highway, in a random lane. Conditions are bright, mostly sunny, and clear.
    
    \item \textbf{Weak OOD setup}: In the weak difficulty, there are two environment configurations. ``Weak high light'' has slightly cloudy and foggy weather, and ``weak noisy low light'' is very cloudy, rainy, and foggy.
    
    \item \textbf{Strong OOD setup}: In the strong setting there are also two environments. ``Strong Low Light'' is at nighttime and is very foggy. Whereas, ``Strong Noisy Low Light'' is more extreme with very rainy, foggy, windy, and dark conditions. 
\end{itemize}

\subsection{Evaluation Metrics}
\subsubsection{Normalized Returns}
The primary metric of evaluation between configurations is the average normalized return, which enables easier cross-domain comparisons. The normalized returns (\ref{eq:norm-ret}) are calculated using the average returns of the agent policy $G_{a}$, a random policy $G_{r}$, and the theoretical maximum return $G_{m}$.

\begin{equation}\label{eq:norm-ret}
    G_{norm} = \frac{G_{a} - G_{r}}{G_{m} - G_{r}}
\end{equation}

\subsubsection{Sample Efficiency}
\cite{surprising} use a qualitative strategy to evaluate sample efficiency, averaging reward curves over numerous runs. Visual differences were used to contrast policy learning speed. In our case, we adopt a threshold approach \cite{smpl-complex-gym}, defining sample efficiency as the number of steps required for the average return to breach 90\% of the max return for that run. A relative threshold helps to disentangle asymptotic model performance and learning speed. For statistical analysis, we also perform permutation testing, using the statistic of the area between the normalized mean learning curve from each model. The null hypothesis is that all curves are sampled from the same distribution. 

\section{Experiments}
For each of our environments, we trained agents using three seeds for each vision model. In the table top task, agents were trained for 1M timesteps of experience, and for the autonomous driving task, 400k timesteps. In the table top task, the seed affects the subset of the YCB objects and positions, and in the autonomous driving task, the seed affects vehicle spawn locations and influences slight weather changes.

\subsection{Sample Efficiency}\label{sec:sample-eff}
\begin{figure}
    \centering
    \includegraphics[width=\columnwidth]{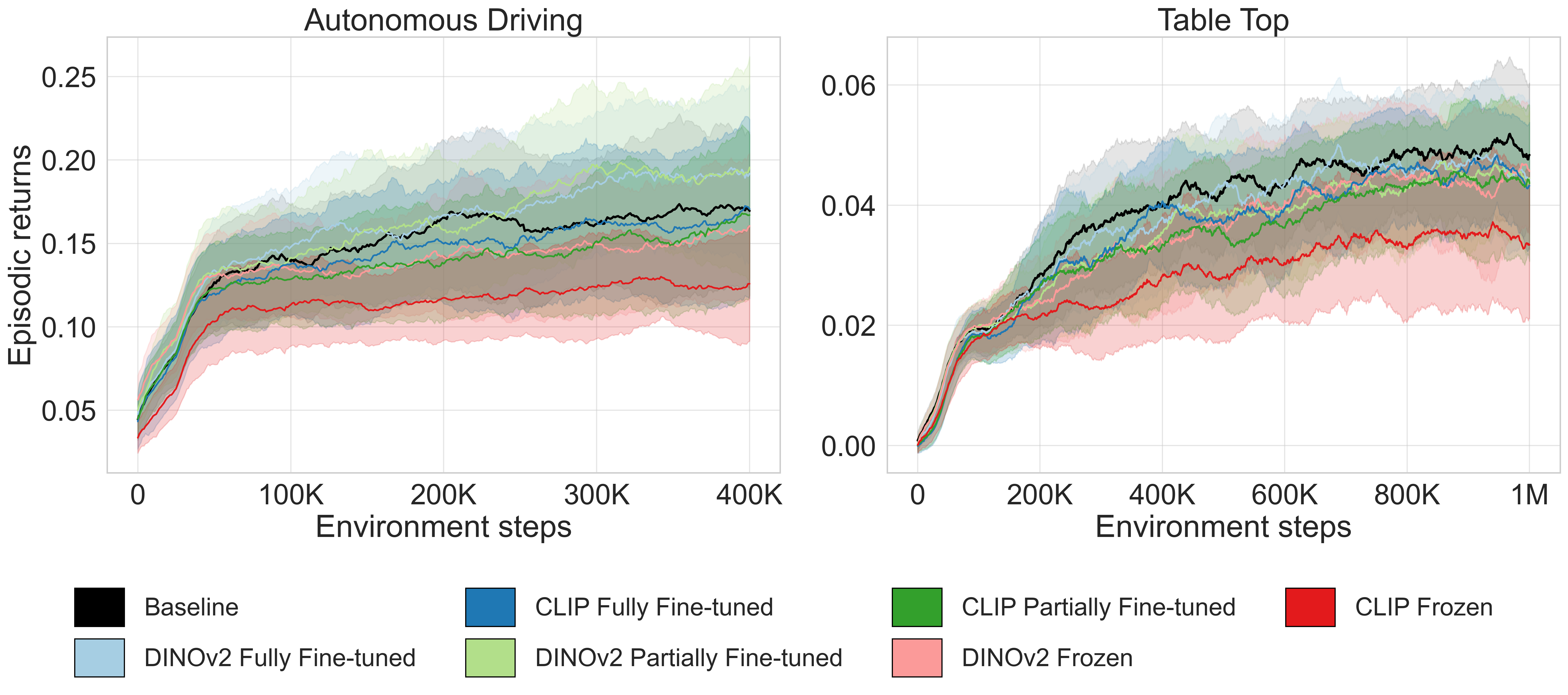}
    \caption{Policy learning curves in both environments. Curves are averaged over three seeds and window-smoothed over 50K steps.}
    \label{fig:curves}
\end{figure}

A primary research question is to ascertain whether end-to-end fine-tuning might result in faster policy learning. Towards this, we start by visualizing the policy training curves for each model and environment. In Fig. \ref{fig:curves}, we average the policy training curves for each agent type across seeds and apply window smoothing over 50k steps. The shaded regions show the standard deviations.

In both environments, there are clear differences in asymptotic performance, but there is no clear visual separation between models in learning rate. In the table top task, the first 200K steps of training are almost identical, and the standard deviation in this region is much tighter than in later stages of training. Regarding asymptotic return, the baseline and most PVM-based agents achieve very similar final average return.

In the autonomous driving environment, again, learning curves are nearly identical for the initial training steps, persisting until approximately 50K steps. However, under increased perceptual and dynamic complexity, the PVMs outperformed the baseline in asymptotic performance. Both DINOv2 policies with fine-tuning attain the highest average return, surpassing the baseline. The CLIP Frozen agent has a significantly lower average return than all other configurations. This finding aligns with the results shown in \cite{clip-not-dense} that off-the-shelf CLIP representations lack dense spatial information. In the autonomous driving task, spatial information is integral for localizing objects or hazards on the road. 

\subsubsection{Understanding World Model Losses}\label{sec:losses}
\begin{figure}[b]
  \includegraphics[width=\columnwidth]{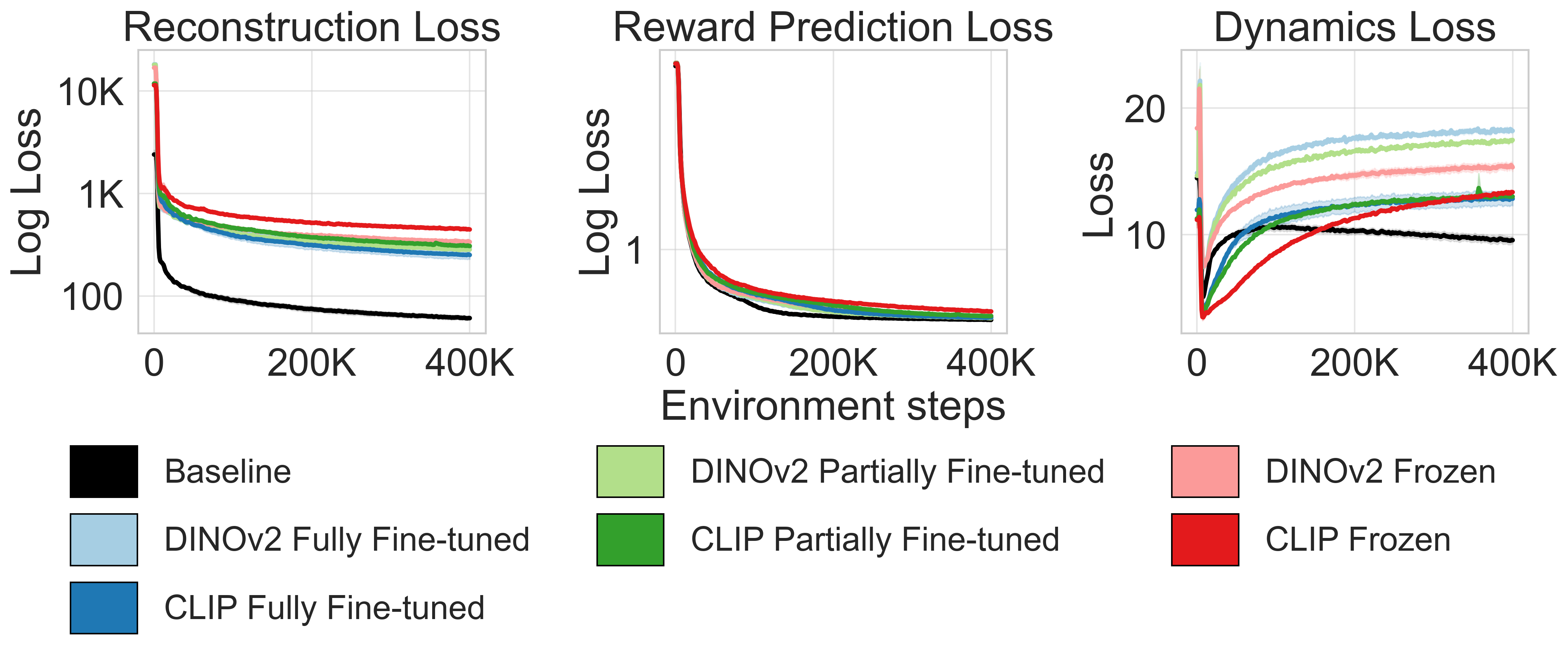}
  \caption{DreamerV3 world model losses in the autonomous driving task, averaged over three seeds and window-smoothed over 2K steps.}
  \label{fig:losses}
\end{figure}
Despite switching the visual encoder in the world model, we do not observe an improvement in the speed of policy learning (Section \ref{sec:sample-eff}). In MBRL, the faithfulness of the world model's imagined trajectories influences policy performance in the real environment. To better understand these findings, we inspect three loss signals in the world model learning: image reconstruction, reward prediction, and dynamics (Fig. \ref{fig:losses}). 

One might expect the image reconstruction loss to stabilize more slowly for the baseline model trained from scratch, particularly in visually complex environments such as the autonomous driving task. In reality, for all vision models, it stabilizes very quickly, well within 25K steps. Additionally, the baseline is able to achieve the smallest loss of all models. The same is seen with reward prediction and indeed most forms of loss in DreamerV3 training, stabilizing within 50k steps for all models. 

In contrast, the dynamics loss took roughly 150K steps to converge for our models. As the policy improves, it explores more of the state space, revealing more complex dynamics that the dynamics model must learn. This is a common cause for the observed V-shaped loss curve \cite{hafner-comment}. One might perceive this challenge as the bottleneck of the world model learning. However, \cite{surprising} showed little correlation between dynamics prediction and ID performance. Indeed, despite a clear separation between models in the dynamics loss starting from $\sim$10K steps, we observe little difference in policy learning rate from this point in Fig. \ref{fig:curves}.

We theorize that the CNN baseline can quickly learn useful image representations because of its higher learning rate than the fine-tuned models, and because CNNs are more sample-efficient than ViTs in general \cite{vit-paper}. The inductive biases of CNNs encourage learning important properties such as locality and translation equivariance, which do not exist in ViTs. As such, ViTs often require many more examples to develop such properties. 

\subsubsection{Quantitative Analysis}
The results of our threshold analysis are shown in Table \ref{tab:combined-thresh}. The range of our values is wide, emphasizing the variability of policy learning even within the same model class. In our results, most ranges overlap, indicating indistinguishable efficiency between our vision models. This is supported by permutation testing (N=10K). Each comparison is done pairwise between every possible combination of visual encoders. In both environments, we find no statistically significant differences between the learning curves. 

\begin{table}[]
\centering
\begin{tabular}{lcc}
& \multicolumn{2}{c}{\textbf{Steps-to-threshold ($10^3$)}}\\
\textbf{Model}  & \textbf{Autonomous Driving} & \textbf{Table Top} \\
\midrule
Baseline                    & 193 (\phantom{0}64 - 217)  & 729 (602 - 787) \\
DINOv2 Fully Fine-tuned     & 265 (260 - 275)            & 601 (504 - 686) \\
DINOv2 Partially Fine-tuned & 298 (284 - 323)            & 618 (450 - 631) \\
DINOv2 Frozen               & 218 (196 - 283)            & 559 (514 - 627) \\
CLIP Fully Fine-tuned       & 236 (180 - 276)            & 750 (693 - 782) \\
CLIP Partially Fine-tuned   & 312 (297 - 314)            & 694 (647 - 780) \\
CLIP Frozen                 & 238 (\phantom{0}56 - 318)  & 559 (420 - 675) \\
\bottomrule
\end{tabular}
\caption{Median steps to reach 90\% of maximum return after window smoothing over 20K steps.}
\label{tab:combined-thresh}
\end{table}

\subsection{OOD generalization}
\begin{figure*}
  \includegraphics[width=\textwidth]{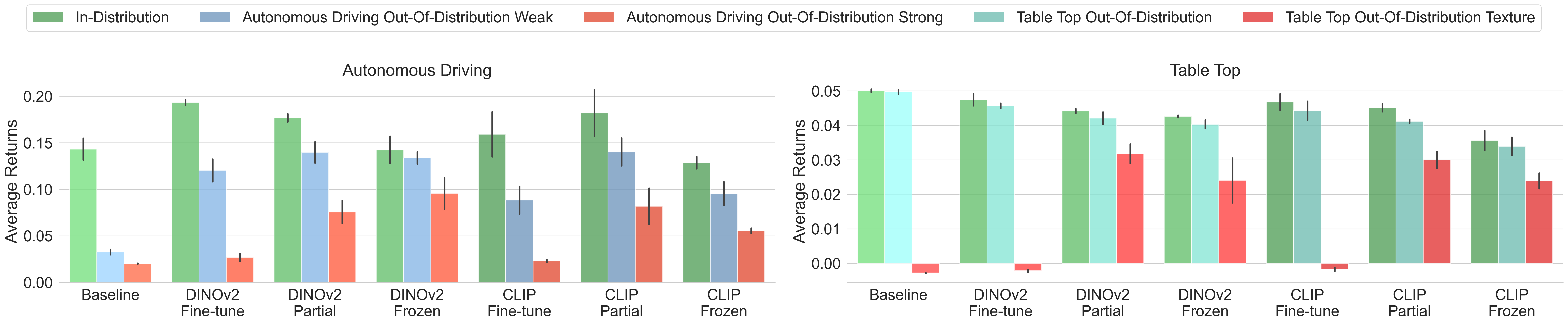}
  \caption{Normalized returns in each environment's evaluation setting for each vision model. Each agent configuration is trained three separate times with different seeds. Black lines represent standard error uncertainty.}
  \label{fig:eval}
\end{figure*}

The trained DreamerV3 agents with different vision encoders are evaluated in simulation in both ID and OOD settings. The results are summarized in Fig. \ref{fig:eval}.

\subsubsection{Table Top Environment}
We evaluate each learned policy on the same 200 episodes for each distribution shift. The baseline encoder achieves the highest ID average return. The next top performers were both agents using PVMs that underwent full end-to-end fine-tuning. Unsurprisingly, we observe a trend where the more fine-tuning we apply to our PVMs, the greater the average ID performance. 

In the original OOD task, we find that PVM-based agents also perform worse than the baseline under modest distribution shifts, replicating the findings of \cite{surprising}. We hypothesize that the domain shift examined in \cite{surprising} favors the baseline CNN architecture. CNNs are naturally translation equivariant, so translating object locations is not a challenging augmentation for the vision encoder. We found that the relative drop in return between ID and OOD settings averages only 4.7\% across all models.

The outcomes are very different under our \textit{hard} distribution shift. The baseline and fully fine-tuned models fail to generalize with the new table texture. We observe the same behavior from these agents, a tendency to ``look away'' from the table. The return of these models is worse than a random agent, a relative drop of 106\% from ID performance for the baseline. Agents using our partially fine-tuned and frozen PVMs maintained a high average return. Partially fine-tuned DINOv2 had the smallest relative drop in performance of 28\%.

\subsubsection{Autonomous Driving Environment}\label{sec:ood-carla}
In the autonomous driving environment, we evaluate agents on ID, weak, and strong distribution shifts. Each difficulty has two environment settings; we average episodic return across the two configurations for each difficulty. We find that the baseline does not achieve the highest ID average return. Strong models include DINOv2, fully and partially fine-tuned, and CLIP, partially fine-tuned. These findings could indicate that in this more challenging task, the larger PVMs can leverage their capacity to outperform the baseline.

We observe a complete collapse in average return for the baseline under both weak and strong visual changes, with relative performance drops of 79\% and 87\%, respectively. Whereas the PVM-based agents do better under \textit{hard} shifts. Both fully fine-tuned models maintain substantial returns under the weak distribution shift, but collapse under the strong shift. The partially fine-tuned and frozen vision models are robust to the severe distribution shifts. In this case, DINOv2 frozen has the smallest relative drop in performance of all models, 6\% and 33\% respectively for weak and strong shifts.

\subsection{Properties of Robust Visual Representations for Control}
\begin{figure}[b]
    \centering
    \includegraphics[width=\columnwidth]{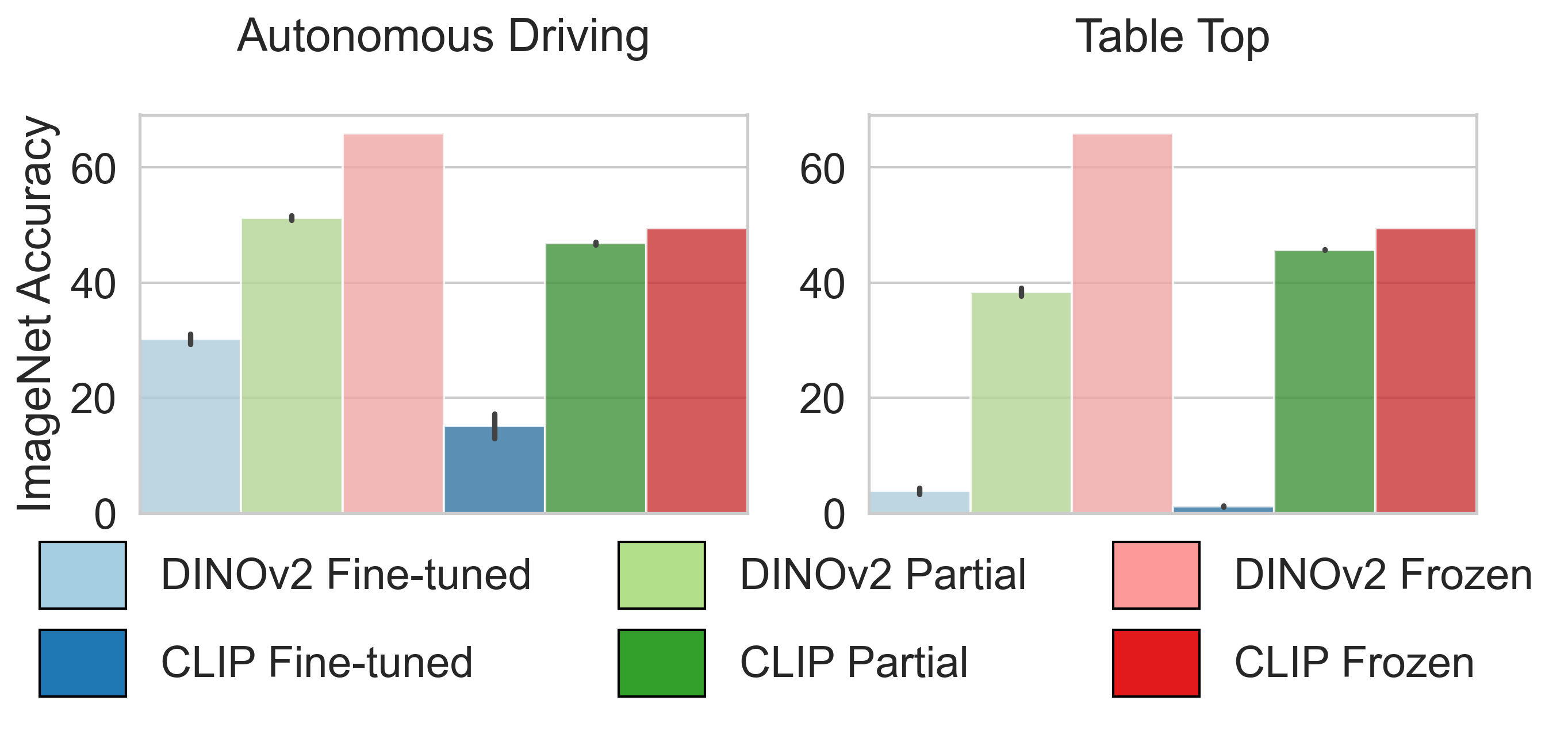}
    \caption{ImageNet accuracy of a K-NN classifier using representations from our PVMs. Values are averaged over model seeds, and black lines represent standard error uncertainty.}
    \label{fig:forget}
\end{figure}

\subsubsection{Visual Invariance with UMAP}\label{sec:invariance}
\begin{figure*}[!htbp]
  \centering
  \begin{subfigure}[b]{0.13\linewidth}
    \includegraphics[width=\linewidth]{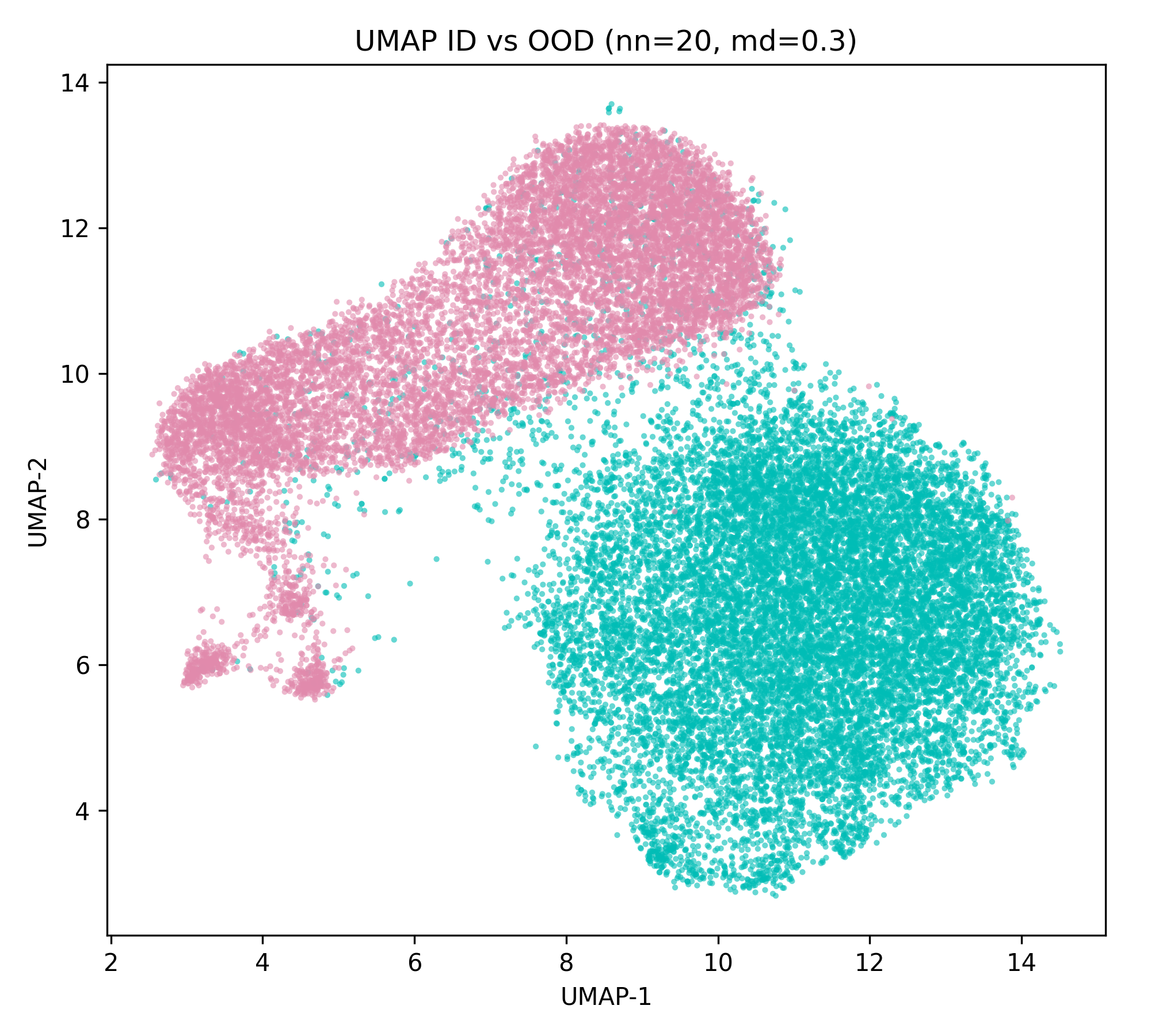}
    \caption*{Baseline}
  \end{subfigure}
  \begin{subfigure}[b]{0.15\linewidth}
    \centering
    \includegraphics[width=0.87\linewidth]{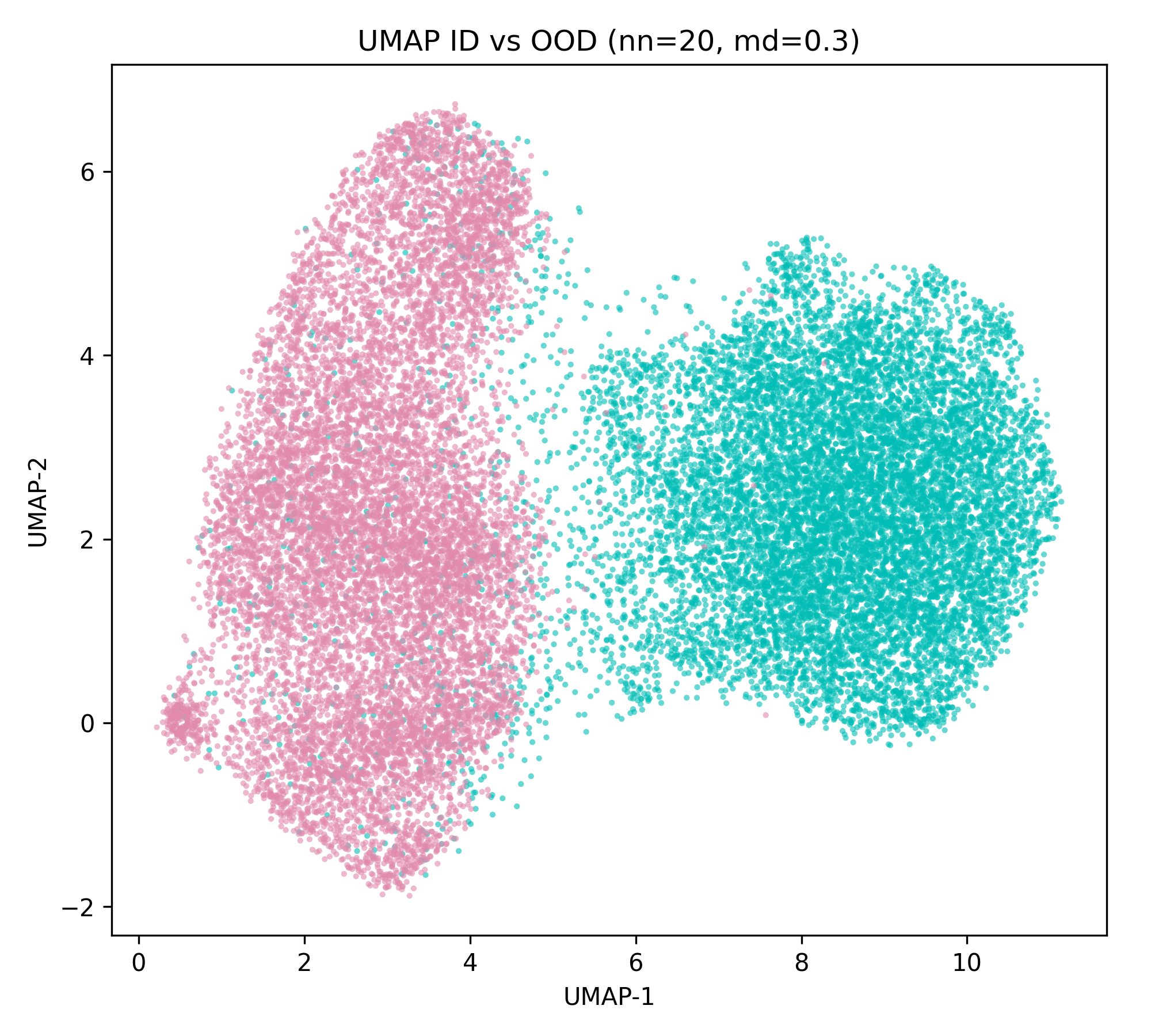}
    \caption*{DINOv2 Fine-tuned}
  \end{subfigure}
  \begin{subfigure}[b]{0.14\linewidth}
    \centering
    \includegraphics[width=0.92\linewidth]{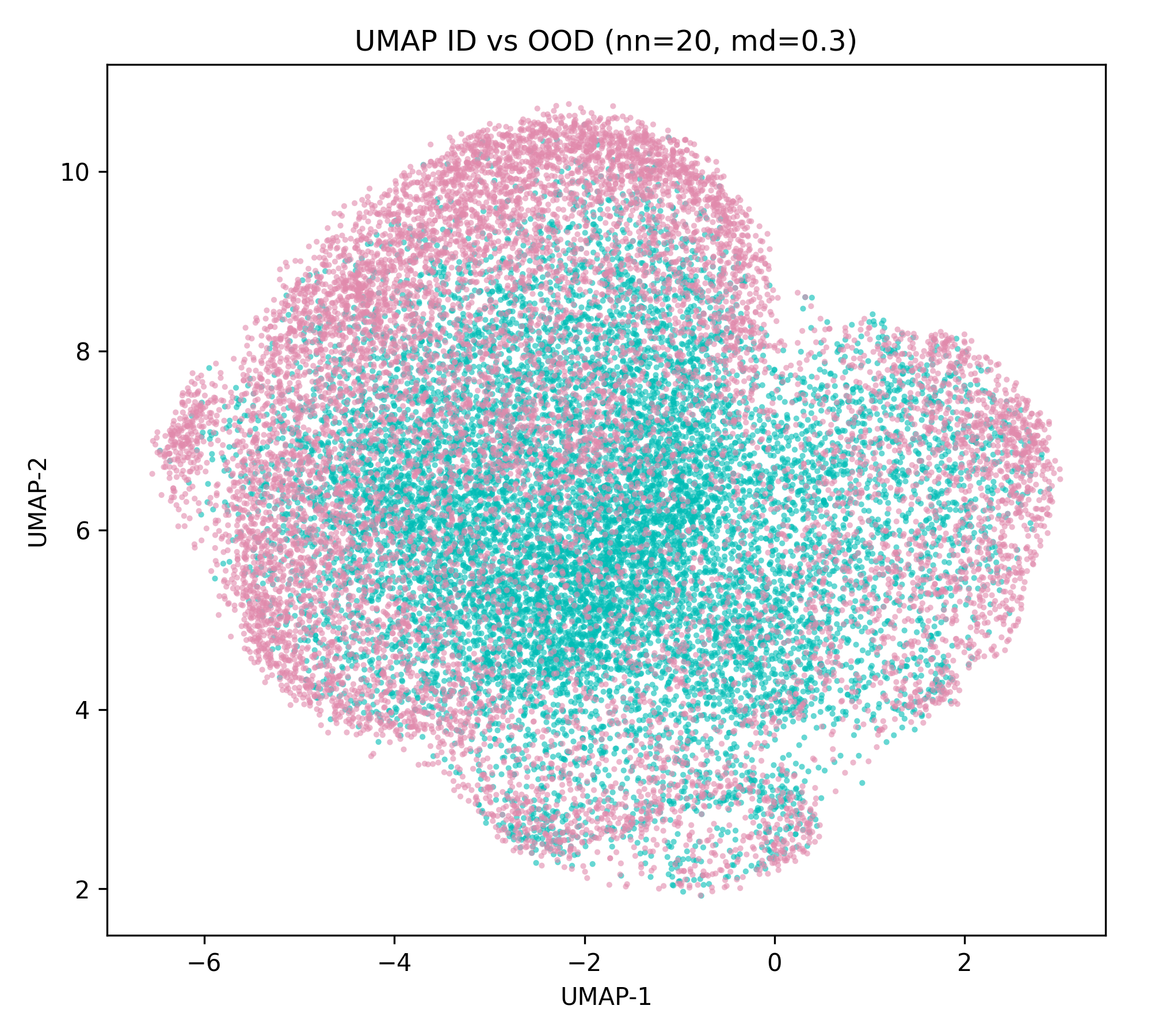}
    \caption*{DINOv2 Partial}
  \end{subfigure}
  \begin{subfigure}[b]{0.14\linewidth}
    \centering
    \includegraphics[width=0.92\linewidth]{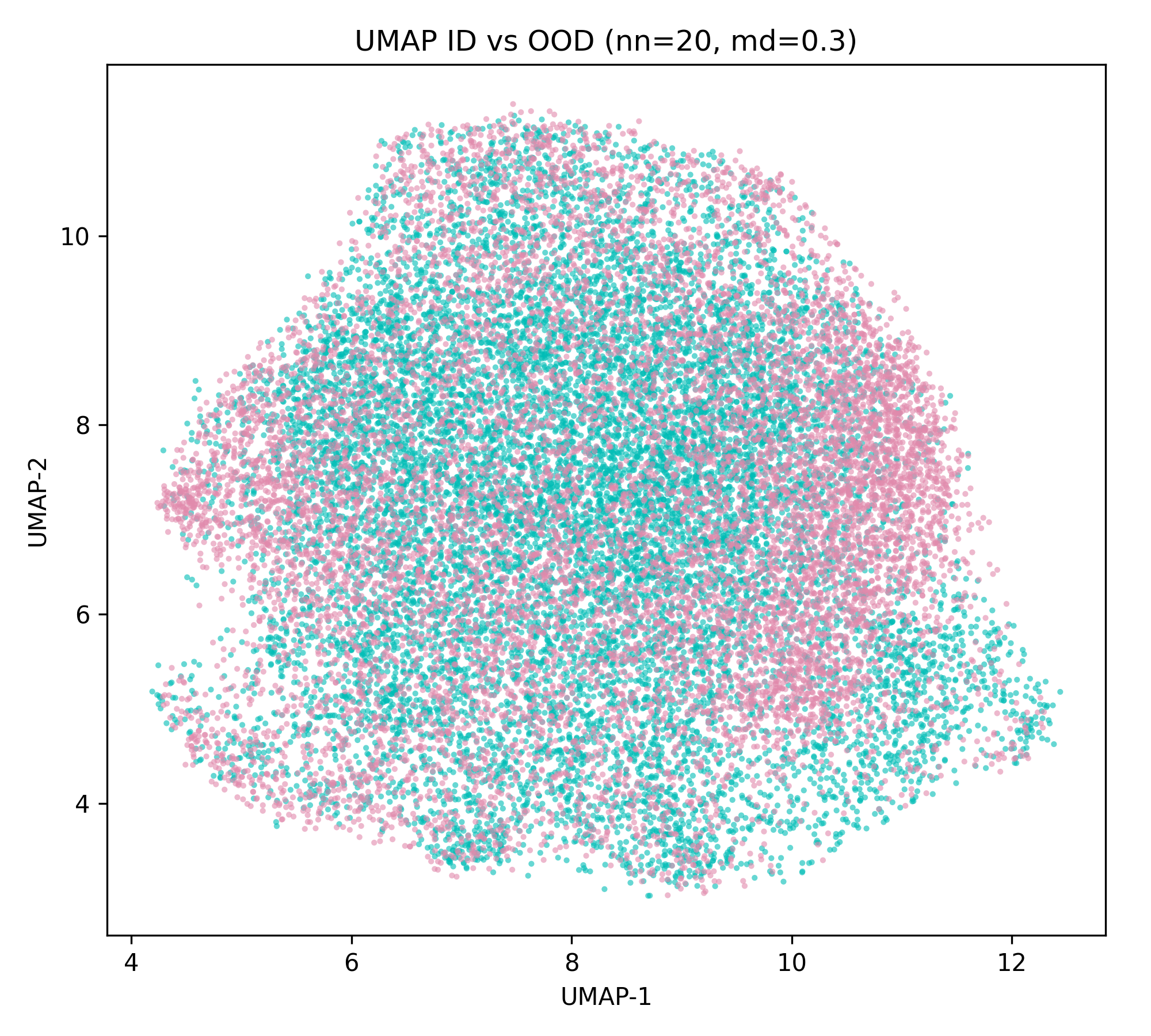}
    \caption*{DINOv2 Frozen}
  \end{subfigure}
  \begin{subfigure}[b]{0.14\linewidth}
    \centering
    \includegraphics[width=0.92\linewidth]{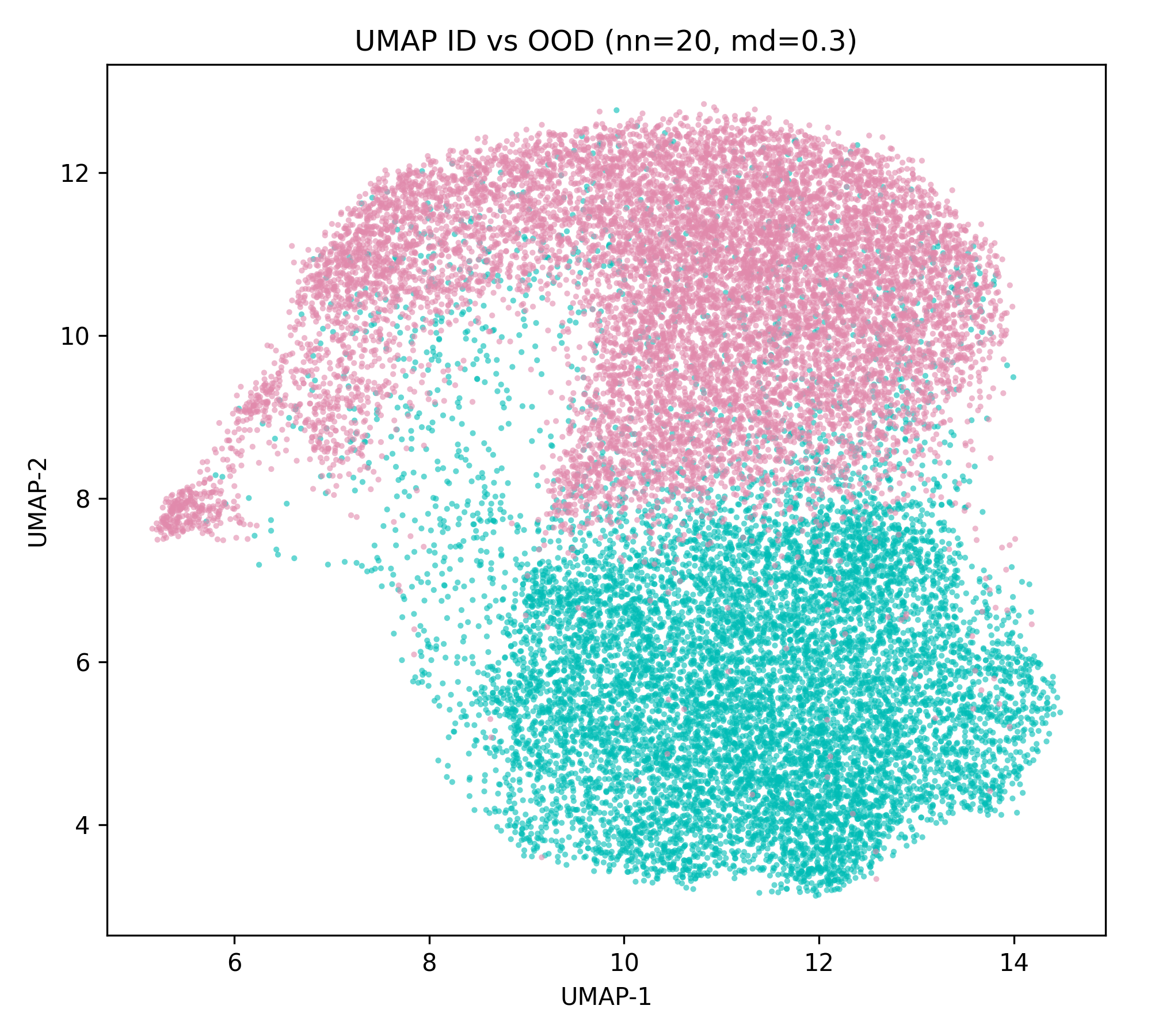}
    \caption*{CLIP Fine-tuned}
  \end{subfigure}
  \begin{subfigure}[b]{0.13\linewidth}
    \includegraphics[width=\linewidth]{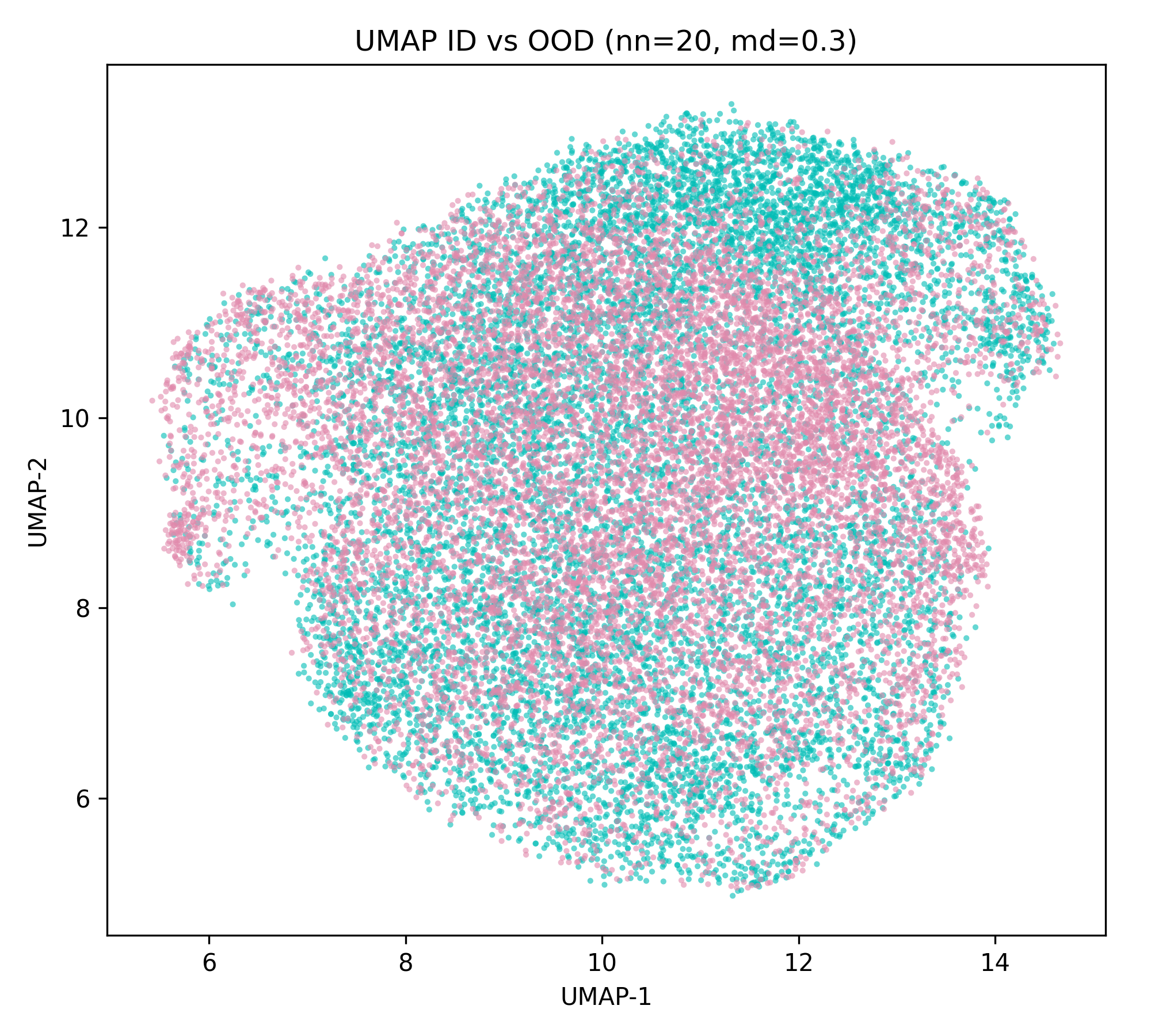}
    \caption*{CLIP Partial}
  \end{subfigure}
  \begin{subfigure}[b]{0.13\linewidth}
    \includegraphics[width=\linewidth]{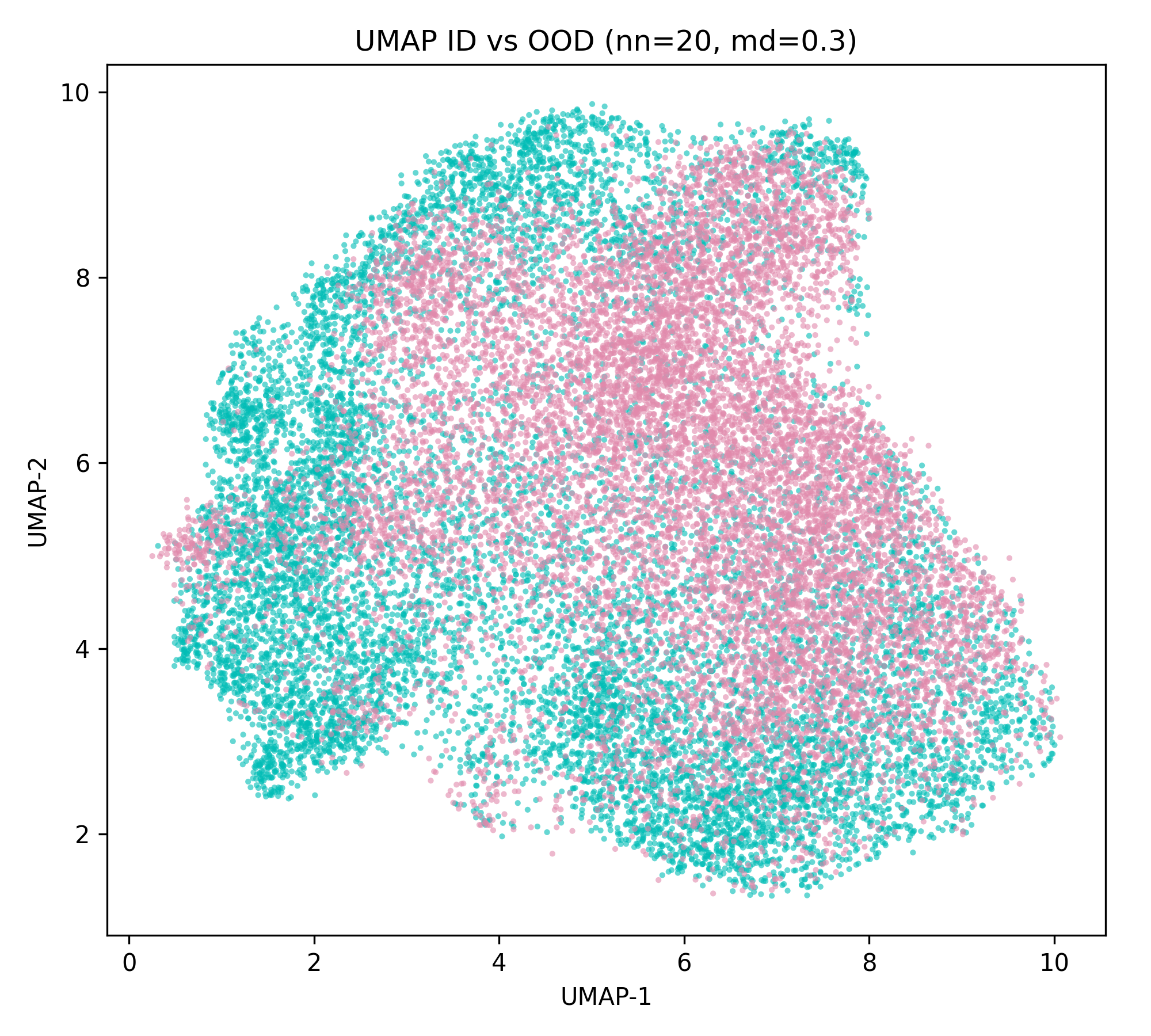}
    \caption*{CLIP Frozen}
  \end{subfigure}
  \caption{Table top UMAP (\texttt{n\_neighbors=20}, \texttt{min\_dist=0.3}) embeddings of DreamerV3 stochastic representations of observations $z_t$ from each of our vision encoders using observations in-distribution \textbf{(blue)} and \textit{hard} out-of-distribution \textbf{(pink)}.}
  \label{fig:umap-comparison}
\end{figure*}

It has been shown that pre-training teaches models invariance to input perturbations, which directly leads to strong OOD generalization \cite{pt-ood}. To represent invariance in our vision models, we use UMAP to project the stochastic DreamerV3 representations $z_t$ to two dimensions (Fig. \ref{fig:umap-comparison}) from ID and \textit{hard} OOD observations.

The results are as expected; the baseline CNN has very distinct separation between ID and OOD representations. With increasing PVM fine-tuning, we observe a gradual loss of visual invariance, and the ID and OOD representations start to form more separate clusters. Though partial fine-tuning still exhibits significant overlap in representations.

Agents using representations with weak visual invariance do not generalize as well. The representations contribute to the world state that is used to inform the actor and critic networks. In such models, performance understandably reduces when evaluated under a distribution shift, as the world state undergoes a drastic change.

\subsubsection{Catastrophic Forgetting}\label{sec:forgetting}

Fine-tuning risks catastrophic forgetting \cite{half-ft-llm}, where a model loses the desirable properties learned in pre-training. To diagnose catastrophic forgetting, we take our PVMs after fine-tuning and use a k-NN classifier (\texttt{n=5}) with their representations for ImageNet-1K classification. For simplicity, we use a small 20K sample of images and an 80/20 training-test split (Fig. \ref{fig:forget}). 

Between our frozen, partially fine-tuned, and fully fine-tuned models, there are significant drops in classification accuracy. The fully fine-tuned models have very poor accuracy, exhibiting severe forgetting. Interestingly, the relative drop in accuracy between frozen and partially fine-tuned is greater between the DINOv2 models. In only one case did a frozen model generalize better than its partially fine-tuned counterpart: DINOv2 in the autonomous driving environment. We theorize that under the most acute distribution shifts, models rely more heavily on invariances learned during pre-training, which, for partially fine-tuned models, only DINOv2 appears to have significantly forgotten. 

\subsubsection{Correlating PVM Properties with Policy Generalization}
Investigations into which properties of frozen PVMs lead to better generalization in visual policy learning \cite{what-makes-pvrs, data-centric}, suggest emergent segmentation ability and ID performance correlate positively with OOD performance, while ImageNet accuracy does not. Our analysis differs as we include models undergoing fine-tuning. We correlate variables with agent performance under the hardest shift for each environment. All values are averaged over the three seeds of each model and are summarized in Table \ref{tab:correlate}. Emergent segmentation ability (or objectness) in ViTs is calculated as the Jaccard index on the VOC dataset, using an interpolated attention map averaged over multiple heads from the last attention block.

We find that the Jaccard index and ImageNet accuracy have strong positive correlations with OOD performance, but ID performance does not. Since we are fine-tuning, the problem we are investigating is different. Both the Jaccard index and ImageNet accuracy can be seen as proxy measures for forgetting. This is why we find generally strong correlations with OOD generalization. Although we see OOD generalization improve from the frozen models to partially fine-tuned models, even though these properties diminish. This indicates that properties, such as segmentation ability, can be sacrificed towards better generalization.

We believe this trade-off is towards boosting policy performance. The partially fine-tuned models can better adjust to the objectives of the DreamerV3 algorithm, resulting in stronger ID performance and lower reward prediction loss. Partial fine-tuning PVMs strikes a good balance between minimal forgetting and strong world modeling.

\begin{table}[b]
\centering
\begin{tabular}{lcc}
& \multicolumn{2}{c}{\textbf{Pearson Correlation}} \\
\textbf{Variable}  & \textbf{Autonomous Driving } & \textbf{Table Top} \\
\midrule
Jaccard index               & 0.80  & 0.81 \\
ImageNet Accuracy           & 0.92  & 0.86 \\
In-distribution Performance & -0.22 & -0.50 \\
\bottomrule
\end{tabular}
\caption{Pearson correlation of different model properties with OOD agent generalization across Autonomous Driving  and table top environments.}
\label{tab:correlate}
\end{table}

\subsubsection{Visualizing Attention Maps}\label{sec:attn}
\begin{figure}[t]
  \centering
    \hfill
    \begin{subfigure}[b]{0.2\columnwidth}
      \includegraphics[width=\columnwidth]{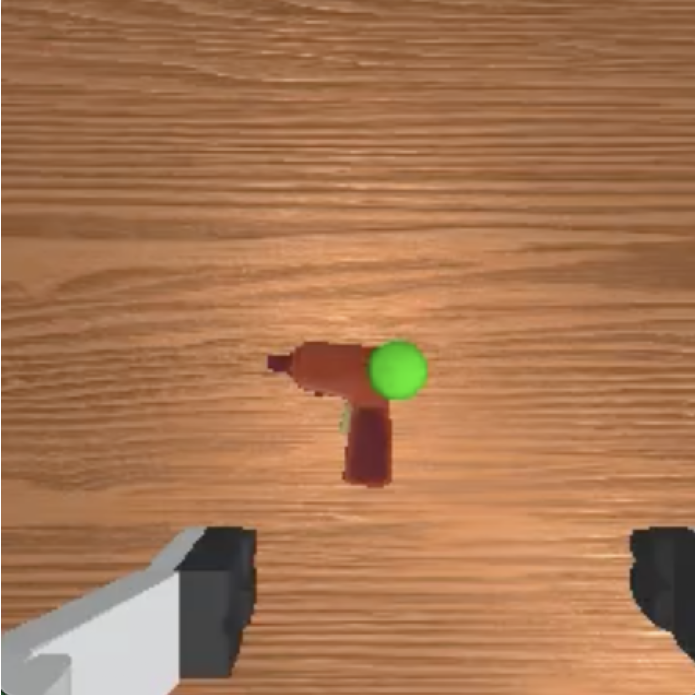}
      \caption*{ID}
    \end{subfigure}
    \hfill
    \begin{subfigure}[b]{0.2\columnwidth}
      \centering
      \includegraphics[width=\columnwidth]{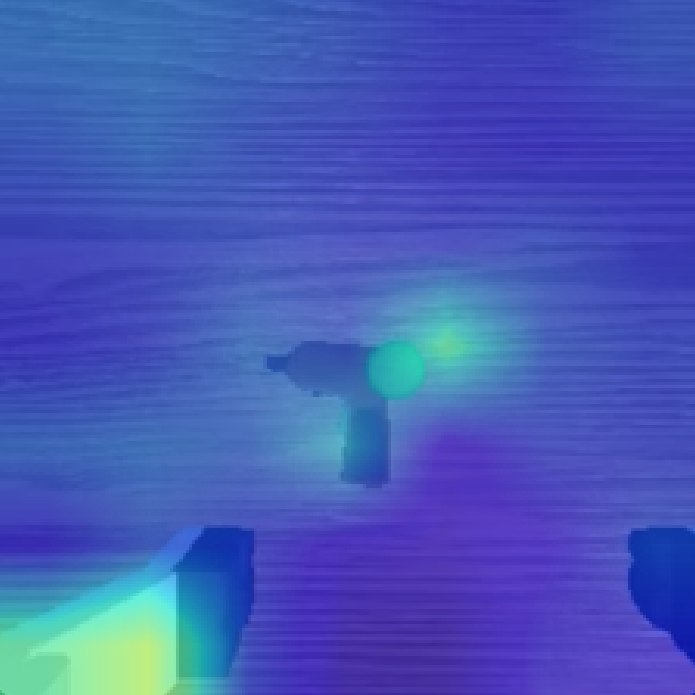}
      \caption*{Fine-tuned}
    \end{subfigure}
    \hfill
    \begin{subfigure}[b]{0.2\columnwidth}
      \centering
      \includegraphics[width=\columnwidth]{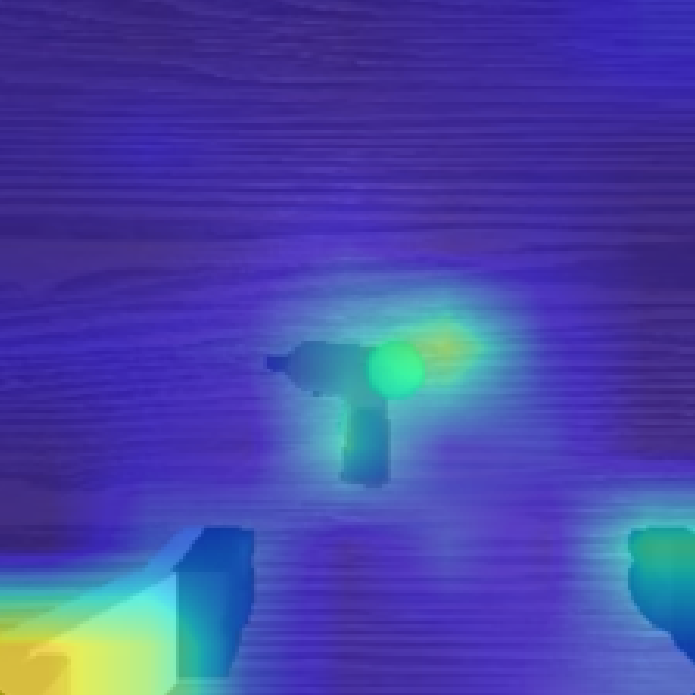}
      \caption*{Partial}
    \end{subfigure}
    \hfill
    \begin{subfigure}[b]{0.2\columnwidth}
      \includegraphics[width=\columnwidth]{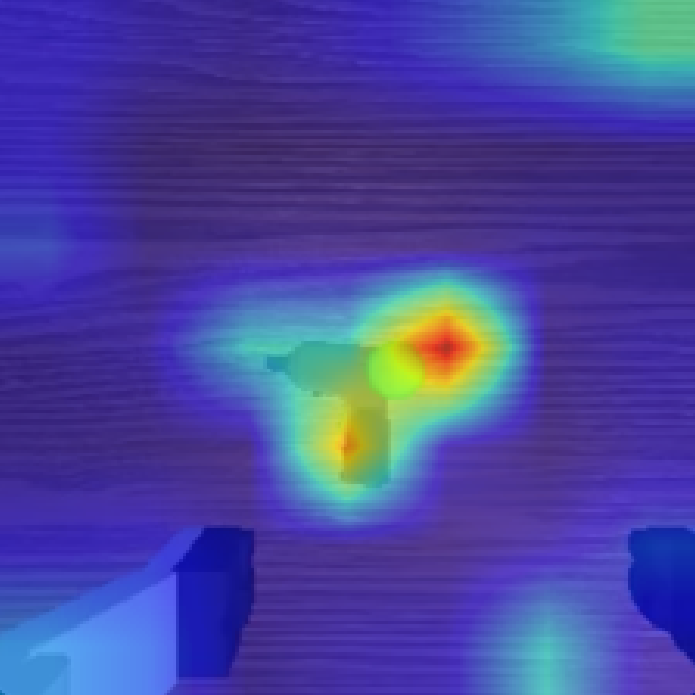}
      \caption*{Frozen}
    \end{subfigure}
    
    \hfill
    \begin{subfigure}[b]{0.2\columnwidth}
      \includegraphics[width=\columnwidth]{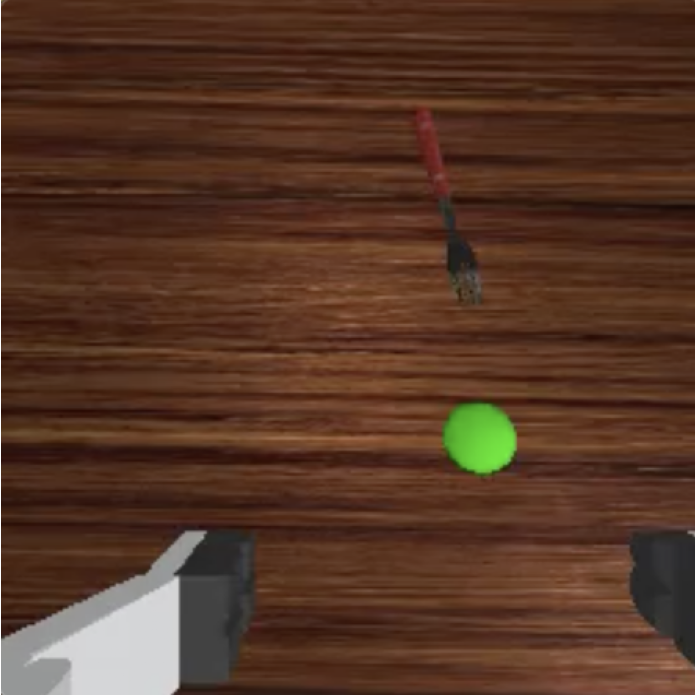}
      \caption*{Hard OOD}
    \end{subfigure}
    \hfill
    \begin{subfigure}[b]{0.2\columnwidth}
      \centering
      \includegraphics[width=\columnwidth]{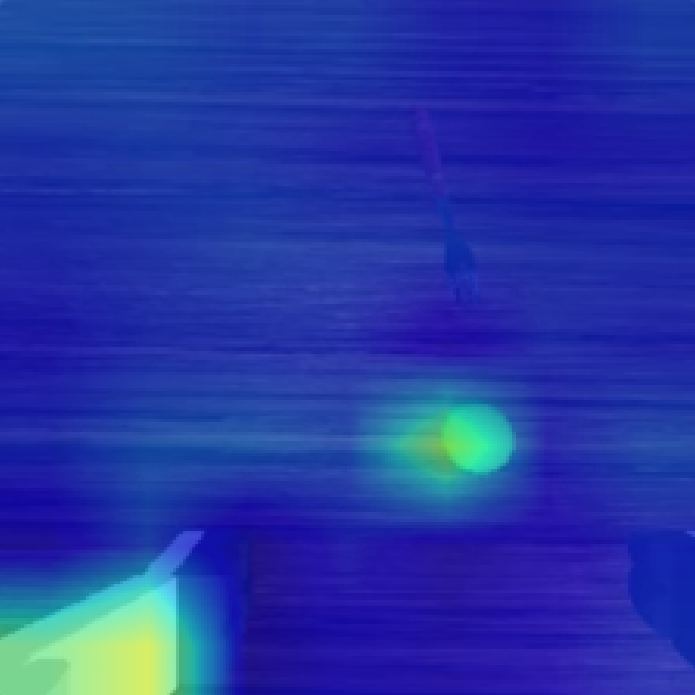}
      \caption*{Fine-tuned}
    \end{subfigure}
    \hfill
    \begin{subfigure}[b]{0.2\columnwidth}
      \centering
      \includegraphics[width=\columnwidth]{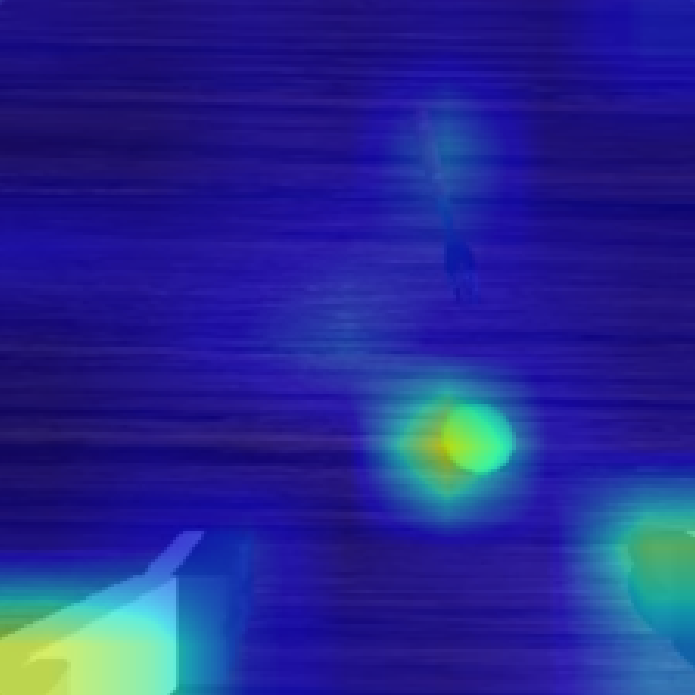}
      \caption*{Partial}
    \end{subfigure}
    \hfill
    \begin{subfigure}[b]{0.2\columnwidth}
      \includegraphics[width=\columnwidth]{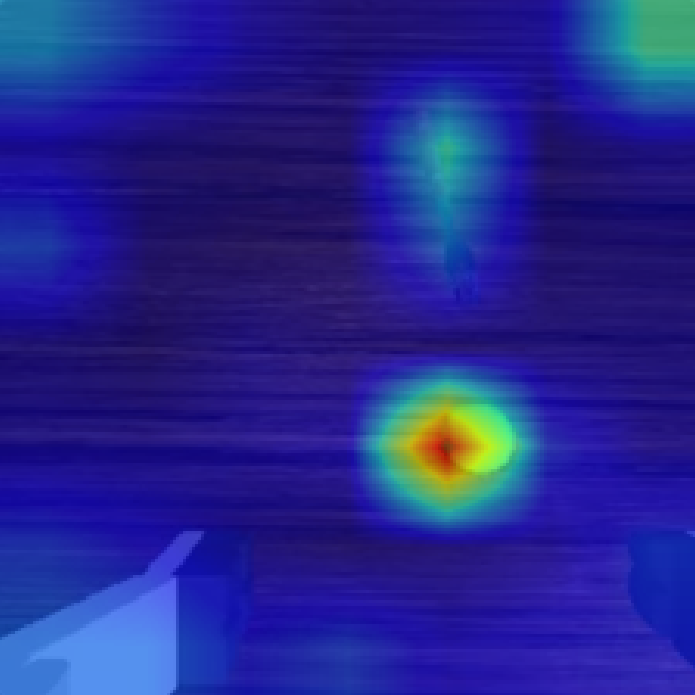}
      \caption*{Frozen}
    \end{subfigure}
    \hfill

  \caption{Attention maps from our CLIP models processing table top observations ID (top) and hard OOD (bottom).}
  \label{fig:clip-attn}
\end{figure}

With our CLIP model, we can visualize the attention map using the \texttt{[CLS]} token to query the attention to each image patch in the last block, averaging across heads (Fig. \ref{fig:clip-attn}). In ID and OOD settings, we observe that all models successfully attend to the goal location, the green sphere. However, only in the ID case do all models attend to the target object, the power drill. In the \textit{hard} distribution shift, only the partially fine-tuned and frozen models have high attention on the fork. Notably, the fully fine-tuned model's attention is more diffuse, attending more to the table. We hypothesize that the model is overfitting and learning to make use of extraneous features of the table's texture to help with the task. It cannot use those cues when the texture is changed, resulting in the ``look away'' behavior we observed.

\subsection{Comparing CLIP and DINOv2}\label{sec:methods-clip-vs-dino}
Comparing both our PVMs, we find that DINOv2 has a stronger segmentation ability compared to CLIP; the Jaccard indices are 0.42 and 0.34, respectively, for our frozen models. This is unsurprising, given CLIP has strong image-level features but is known to be less suitable for dense prediction  \cite{clip-not-dense}. Comparing across the same level of fine-tuning, both ID and OOD returns are similar. Only CLIP frozen has much worse performance than DINOv2 frozen. This finding aligns with prior research indicating that CLIP embeddings can become effective for spatial tasks through fine-tuning \cite{clip-not-dense}. 

\section{CONCLUSION}
In this work, we show that the use of PVMs in MBRL can be an effective approach to improve generalization. In both environments tested, we observe a drastic collapse in the performance of the CNN baseline once exposed to \textit{hard} distribution shifts. Full fine-tuning, while improving ID performance, did not seem to benefit generalization. Partially fine-tuned and frozen vision models both exhibited the strongest generalization performance. Considering CLIP and DINOv2 separately, in three out of four cases, agents using partial fine-tuned PVMs had the greatest average return under hardest distribution shifts. Consistent with the findings of \cite{surprising}, our results show no improvement in terms of sample efficiency. While \cite{surprising} attribute the training speed limitation to the fixed representation of their frozen models, we show that even end-to-end fine-tuning representations do not impact efficiency. We instead propose that it is the inductive biases of the baseline CNN that enable it to quickly learn effective representations. While our study focused on a relatively small set of environments and models, it provides valuable insights and a solid foundation for further investigation. Future work will build on this foundation by incorporating additional model-based algorithms, recent PVMs, and a wider variety of robotic tasks, further strengthening the robustness and generality of the findings.








\bibliographystyle{bib/IEEEtran}
\bibliography{root}

\end{document}